\documentclass[
    twocolumn,
    hf,
]{ceurart}
\usepackage[top=2.5cm, bottom=2.5cm, left=2.5cm, right=2.5cm]{geometry}

\usepackage[frozencache,cachedir=./_minted-fail]{minted}
\setminted{breaklines=true}
\usepackage[utf8]{inputenc}	
\usepackage{placeins} 
\usepackage{lipsum}
\usepackage{tabularx}
\usepackage{metrix}

\begin{document}

\copyrightyear{2022}
\copyrightclause{Copyright for this paper by its authors.
    Use permitted under Creative Commons License Attribution 4.0
    International (CC BY 4.0).}

\conference{}
\title{(Not) Understanding Latin Poetic Style with Deep Learning}


\author[1]{Ben Nagy}[%
    orcid=0000-0002-5214-7595,
    url=https://github.com/bnagy/fail-paper,
    email=benjamin.nagy@ijp.pan.pl,
]
\address[1]{Institute of Polish Language, Polish Academy of Sciences (IJP PAN)\\
    Adama Mickiewicz 31\\
    Kraków, Poland}
\begin{abstract}
    This article summarizes some mostly unsuccessful attempts to understand
    authorial style by examining the attention of various neural networks (LSTMs
    and CNNs) trained on a corpus of classical Latin verse that has been encoded
    to include sonic and metrical features. Carefully configured neural networks
    are shown to be extremely strong authorship classifiers, so it is hoped that
    they might therefore teach `traditional' readers something about how the
    authors differ in style. Sadly their reasoning is, so far, inscrutable.
    While the overall goal has not yet been reached, this work reports some
    useful findings in terms of effective ways to encode and embed verse, the
    relative strengths and weaknesses of the neural network families, and useful
    (and not so useful) techniques for designing and inspecting NN models in
    this domain. This article suggests that, for poetry, CNNs are better choices
    than LSTMs---they train more quickly, have equivalent accuracy, and
    (potentially) offer better interpretability. Based on a great deal of
    experimentation, it also suggests that simple, trainable embeddings are more
    effective than domain-specific schemes, and stresses the importance of
    techniques to reduce overfitting, like dropout and batch normalization.
\end{abstract}


\maketitle

\begin{figure*}
    \includegraphics[width=\linewidth]{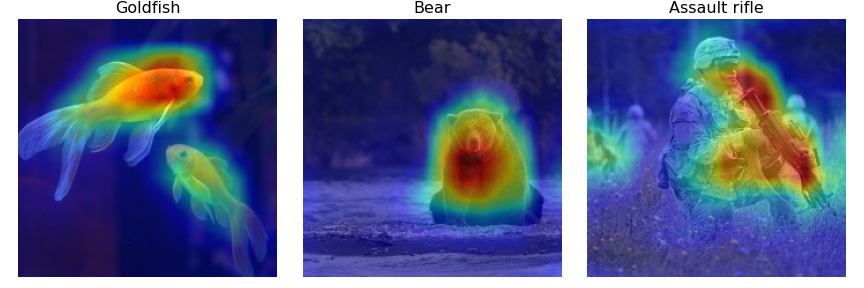}
    \caption{An example from the \mintinline{text}{tf-keras-vis} documentation
        of \mintinline{python}{ScoreCam} working as intended, on images.}
    \label{fig:sc_img}
\end{figure*}

\section{Introduction}

Failing to understand authorial style in Latin verse will, at first glance,
hardly seem novel---generations of students, including this author, have
participated enthusiastically in the field. The major contribution of the
present article, however, is to bring to the endeavor a selection of techniques
from the modern domain of \emph{neural networks} (NNs). Carefully configured NNs
are shown to be extremely strong authorship classifiers, and it is hoped that
they can one day teach `traditional' literary critics something interesting
about how the authors differ in style. Regrettably, at least at this stage, the
ways in which the neural networks reach their decisions seem even more difficult
to understand than the poets themselves.

At the outset, it should be stated that this article is focused only on poetry,
not prose. A great deal of computational stylometry, as well as a great deal of
work on neural networks has already been done on texts whose basic atomic
elements are words, or characters. For poetry, something different is needed.
Sonic effects like rhyme, alliteration, or assonance are often activated between
word boundaries, and metrical constraints like stress (and, in Latin, like
quantity) are tied to the syllable. Working at the level of syllables, two
metaphors suggest themselves: music, and images. To think about a poem
musically, we treat it as a rhythmic sequence of syllables, where words and
phrases are `melodies', punctuated naturally by regular recurring elements like
caesurae and line-breaks. For this metaphor, Recurrent Neural Networks (RNNs, of
which the most commonly used these days are LSTMs) are a good fit. However if we
think of a poem visually, as a `sound picture', then there is relevance in
\emph{adjacency}---rhyme becomes a `vertical line' of syllables with vowels of
the same `colour'. An extended string of sonorous sounds and lilting, liquid
onsets should `look' different to staccato collections of clattering consonants.
This is where Convolutional Neural Networks (CNNs), currently the tool of choice
for image classification, should be ideal. At least in theory.

This article restricts itself to classical Latin verse, a quantitative tradition
mostly adapted from Greek dactylic hexameter. Apart from practical concerns
(Latin verse being my area of interest, expertise and research), Latin verse has
both strengths and weaknesses with respect to the metaphorical frameworks just
discussed. It has been shown elsewhere \cite{nagy2021metre} that Latin hexameter
is stylistically dense, and that metrical features (the `rhythm' in our musical
metaphor) are, in their own right, privileged indicators of authorial style. In
fact it has been suggested that metre has a `semantic halo'
cross-linguistically, imparting meaning independent of content
\cite{sela_semhalo}. The case is slightly harder to make for the `sound
picture'. Classical verse does not follow a regular rhyme scheme. But, although
some scholars still deny any deliberate use of rhyme, there is plenty of
`traditional' scholarship
\cite{herescu_poesie_1960}\cite{deutsch_1978}\cite{clarke_intentional_1972}\cite{guggenheimer_rhyme_1972}
(not an exhaustive list) as well as quantitative stylometry
\cite{nagy_rhyme_2022} arguing that rhyme was an element of style well before
the fourth century \textsc{ce} which is commonly taken as its inception. At the
least I expect no expert would deny that sonic effects of one kind or another
play a part in classical style. As to the broader applicability of this
research, one anticipates that it would be much the same everywhere---some
traditions are better suited to the `sound picture', others better to the `music
of metre', but hopefully all, at least to some extent, to both.

\section{Poetic Corpus}

As a basis for this initial investigation, a moderately sized corpus of Latin
hexameter was chosen for examination (Table \ref{tab:corpus}).%
\footnote{ The corpus is based on the digital editions from MQDQ
    \cite{mqdq_2007}, post-processed using software previously developed by the
    author \cite{nagy_mqdq_2019}.}
One of the goals for this research was to see whether NNs could identify
vertical patterns (whether sonic, metrical or simply lexical), and so hexameter
verse, with its continuous and homogenous metrical scheme, was the easiest place
to start. The corpus is not well balanced in terms of size-per-class, and
contains only one author with multiple works (the \emph{Georgics} and
\emph{Aeneid} of Vergil), so `authorship classification' is mostly, in this
instance, `work classification'. This was not by design, but is simply an
unfortunate fact of life---most authors produced only one or two long pieces in
this metre.

\begin{table}
    \caption{A summary of the test corpus.}
    \label{tab:corpus}
    \centering
    \begin{tabularx}{\linewidth}{XXr}
        Author           & Work                   & Lines  \\
        \midrule
        Ovid             & \emph{Metmorphoses}    & 12009  \\
        Vergil           & \emph{Aeneid}          & 9840   \\
        Vergil           & \emph{Georgics}        & 2188   \\
        Horace           & \emph{Satires}         & 2113   \\
        Juvenal          & \emph{Satires}         & 3832   \\
        Lucretius        & \emph{De Rerum Natura} & 7375   \\
        Statius          & \emph{Thebaid}         & 9744   \\
        Valerius Flaccus & \emph{Argonautica}     & 5561   \\
        Lucan            & \emph{Pharsalia}       & 8060   \\
        Silius Italicus  & \emph{Punica}          & 12200  \\
        \midrule
                         & Total Lines            & 72922  \\
                         & Total Words            & 470377 \\
        \midrule
    \end{tabularx}
\end{table}

\section{Methods}

\subsection{Care and Feeding of Neural Networks}

All NN experiments were run in Python using Tensorflow
\cite{tensorflow2015-whitepaper} with Keras \cite{chollet2015keras}, running on
Linux with NVIDIA GPUs (using CUDA). This is mentioned with regard to
replication, since some inconsistencies were observed when running the models on
other GPUs (like the Apple M1 Neural Engine). Using the model configurations in
the Appendix, similar models could no doubt be built in different frameworks.

\subsubsection{Encoding and Embedding}\label{sec:embed}

Before the application of computational techniques, data must be encoded. From
the outset it was clear that sub-word tokenisation would be required, since much
of the metrical metadata (stress, length, etc.) occurs at the level of syllables
rather than words. Tokenizing this way would also (hopefully) make it more
difficult for word-level overfitting to occur. In the initial phases of the
research, tokenization was applied at the metron level, using a hand-designed
encoding scheme with nine channels per metron to encode Onset, Nucleus, Coda,
Length, Strong Caesura, Weak Caesura, Diaeresis, Ictus/Accent Conflict,%
\footnote{ This is a technical consideration in Latin philology. Briefly, there
    is a stylistic component to the interplay between the primary stress of a
    word (the \emph{accent}) and the beginning of a metrical foot (the
    \emph{ictus}). For the  gory details, an interested reader might begin with
    the classic introduction by Winbolt \cite{winbolt_latin_1903}}
and Elision. The final six of those channels were simple binary values, and the
three channels for phonetic data were continuous variables in $[0,1]$ arranged
phonologically from `front' to `back'. The encoding was applied at the level of
metrons because a regular line of dactylic hexameter always contains exactly
twelve metrons, and so vertical effects would be preserved. This scheme is
mentioned mainly as a cautionary tale because it performed noticeably worse than
the more basic scheme that follows using syllable-level encodings with a
trainable embedding layer.

\begin{figure}[h]
    \footnotesize
    \begin{verbatim}
Ecce Lichan trepidum latitantem rupe cauata
Adspicit, utque dolor rabiem conlegerat omnem,
"Tune, Licha", dixit "feralia dona dedisti?

ekke Likan trepidum latitantem rupe kawata
adspikit utkwe dolor rabiem konlegerat omnem
tune Lika diksit feralia dona dedisti

ek+A+L+S ke+WC li+S kan+A+L+SC tre+S pi
dum+A+L+SC la ti tan+A+L+S tem+L+DI ru+A+L+S
pe+WC ka wa+A+L+S ta+L EOL ad+A+L+S
spi kit+DI ut+A+L+S kwe+WC do+S lor+A+L+SC
ra+S bi em+A+L+SC kon+L le+A+L+S ge ...
\end{verbatim}
    \normalsize
    \caption{The stages of transformation from raw verse, to phonetic orthography,
        to syllable tokens with metrical metadata.}
    \label{fig:syl_transform}
\end{figure}

\begin{figure}
    \includegraphics[width=\linewidth]{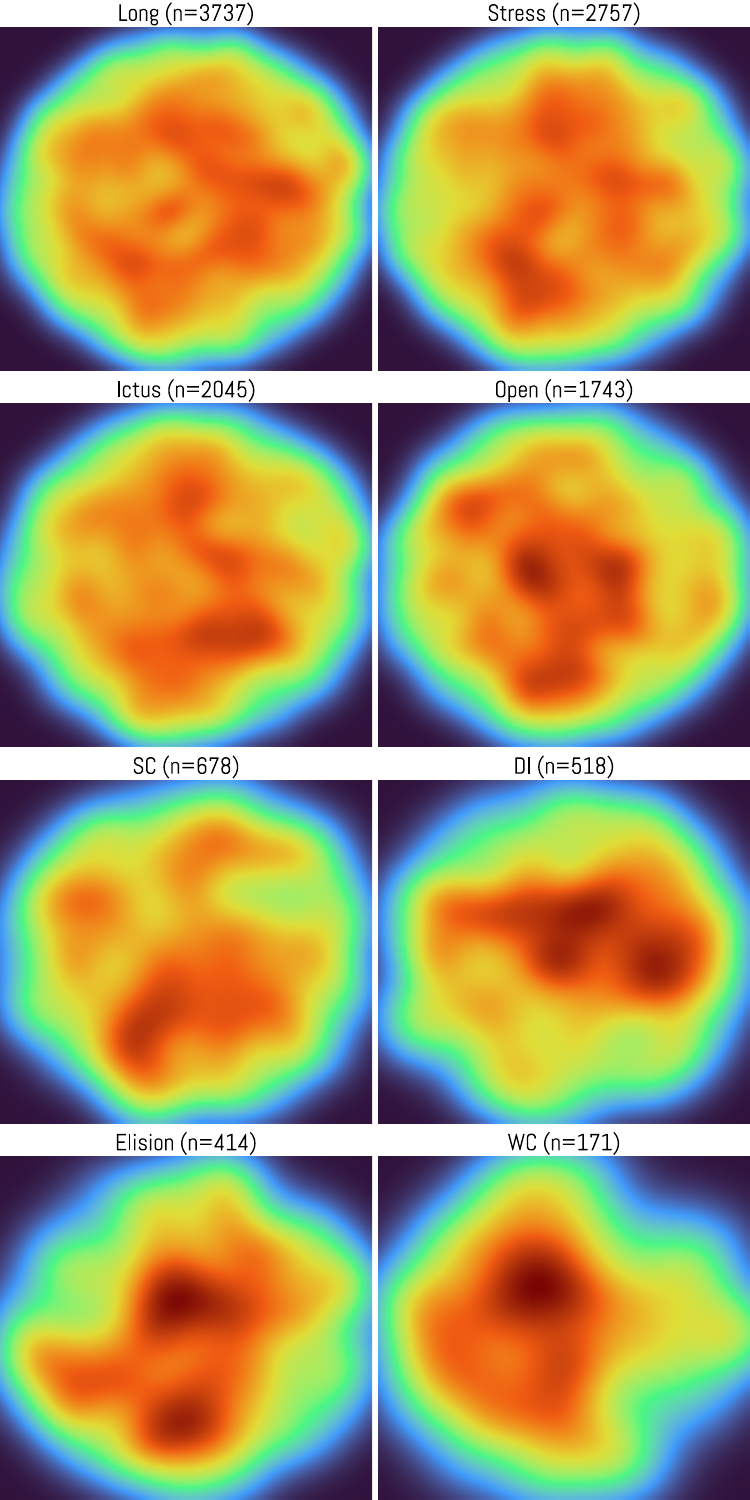}
    \caption{Do the embeddings carry meaningful poetic information? The
        embedding space is projected to 2D using UMAP, and then density
        estimates are plotted for syllables with certain poetic features to
        determine whether they cluster.}
    \label{fig:emb_dense}
\end{figure}

In the second attempt, the data was tokenized per-syllable, as opposed to
per-metron. This of course yielded lines of uneven length,%
\footnote{By convention, the Latin hexameter line (almost) always ends with a
    dactyl and a spondee (\metricsymbols{_ uu , _ _}), but the first four feet may be
    either, allowing between 13 and 17 syllables per line.}
which required slightly more complicated padding when CNNs were used. Syllables
were transformed to a phonetic representation,%
\footnote{ The phonetic encoding process is described in full in previous work
    \cite{nagy_rhyme_2022}.}
and then augmented with metrical metadata (Fig. \ref{fig:syl_transform}). The
final tokens were then simply encoded (as sequential integers, not one-hot),
yielding a lexicon with about ten thousand unique tokens. The benefit of this
approach is that the embedding layer could be made trainable and moved into the
model. This allowed for custom syllable embeddings to be produced and then
frozen (made non-trainable) and re-used for future models. Models trained from
scratch---but using the pre-trained embeddings---are exponentially quicker to
train, and usually more accurate, so it is clear that there are some sort of
emergent stylistic properties in the trained syllable embeddings, much as word
embeddings will evolve to reflect an underlying semantic topology. But can we
learn anything from them? In Figure \ref{fig:emb_dense},%
syllables were separated according to certain metrical features (all syllables
followed by a strong caesura, all long syllables, etc.) and then those syllables
were isolated in a 2D projection using UMAP \cite{mcinnes_umap_2018} to see if
certain kinds of syllables were grouped together in the embedding space. Broad
phonological features, represented by the classes Long (vowel quantity), Accent
(syllable stress) and Open (syllables without final consonants) did not appear
to be captured. There is some indication that word-final syllables (SC, WC, DI)
\emph{are} weakly clustered according to what philologists classify as different
kinds of pauses%
\footnote{In Latin dactylic verse, the pauses are classified according to where
    the end-of-word occurs in the metrical foot. When the end of the word occurs
    at the end of a foot it is a \emph{diaeresis} (DI), after the initial longum
    of a dactyl or spondee a \emph{strong caesura} (SC) and a \emph{weak
        caesura} (WC) after the first breve in a dactyl.}
which is interesting, but it must be noted that the absolute numbers of such
syllables are much lower than the other categories observed, so the density
estimates may be over-sensitive.

\subsubsection{Controlling Overfitting}

Neural Networks in general are well-known to overfit (particularly with small
datasets like this), and it can be challenging to train models that generalise
well. If a model overfits too quickly it will never learn to generalise, because
the loss gradient drops to zero, preventing further learning. While
experimenting with hundreds of different model designs the general trend that
emerged was that the standard techniques of \emph{dropout} and \emph{batch
    normalization} were effective and indispensible tools. Models were found to
tolerate surprisingly high levels of dropout at multiple points. For this task,
batch normalization was found to be more effective when applied before dropout,
not after as is sometimes recommended. The other standard approach that worked
well was \emph{parameter starvation}. In this technique, models are tuned down
to reduce the total number of parameters. In CNN models this can be done by
reducing the number of convolutional layers as well as the number of filters in
those layers. In RNNs, the number of cells per layer can be reduced. In both
types of models the fully-connected (\emph{dense}) layer(s) preceding the final
\emph{softmax} (classification) layer can be reduced---there is no need for the
dense layers to match the final width of the convolutional sequence. For these
experiments, models were starved until they began to underfit on the training
data, and then parameters were slightly increased. This led to models that, in
deep learning terms, are tiny. The final CNN model (\ref{app:cnn}) has 572k
trainable parameters, the LSTM model (\ref{app:lstm}) 338k. To compare, the
well-known image classification model VGG16 \cite{vgg16} has about 138 million
parameters, and is `small' in comparison to billion-scale monsters like GPT-3 or
DALL-E.

\section{Results}

\subsection{Classification Results}

It has been shown elsewhere that Latin hexameter verse is stylistically dense
\cite{nagy2021metre} and is well classified by most supervised machine-learning
methods, so there is no need to turn to Deep Learning for an additional
percentage point or two of accuracy. The main intent of this section is simply
to compare the NN approaches to each other, and to note that (particularly with
good embeddings) almost any sensible NN will work. To report properly on
accuracy would require dozens of train-test cycles, which particularly for LSTMs
is quite computationally expensive---a single training cycle for the LSTM model
(\ref{app:lstm}) is around two hours on an NVIDIA GeForce RTX 3090. By contrast,
various CNN models, using pre-trained embeddings, train to a similar accuracy in
roughly one minute: around 97--98\% (simple accuracy, 64-line samples). But
frankly, the precise accuracy figures are a secondary concern---we are not here
to build classifiers, we are here to try to interpret models. As with previous
research, it is still clear that there is `plenty of style to go around'. The
same models trained only on metrical metadata (removing all phonetic/lexical
data) already reach an accuracy of about 93\%, comparing favourably with
previous work \cite{nagy2021metre}. When removing the metrical component instead
(using only phonetic syllable data), the figure is roughly the same; around
92\%. Authorial style in this genre is extremely rich, pervading both words and
metrical expression. Given the hugely improved training speed and equivalent
accuracy, CNNs are presumptively preferred at this point, if all other factors
are equal. Next, we examine the interpretability of the two approaches.

\begin{figure*}
    \includegraphics[width=\textwidth]{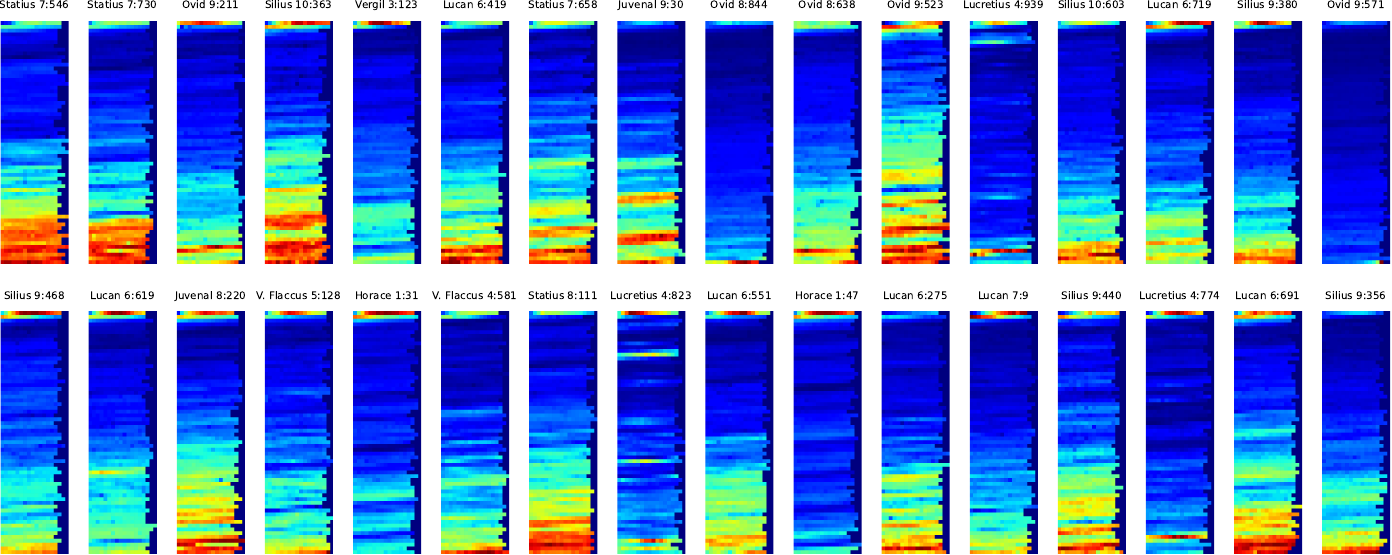}
    \caption{ LSTM attention from the \mintinline{python}{VanillaSaliency}
        visualizer, on 64-line chunks. The \emph{vanishing gradient} problem is
        clearly visible.}
    \label{fig:thumbs32}
\end{figure*}

\subsection{Visualising Attention}

The most promising aspect of these methods is the ability for us to examine the
\emph{attention} of the NNs. Experiments were conducted with three visualizers
implemented in the recent \mintinline{python}{tf-keras-vis} package
\cite{Kubota_tf-keras-vis_2021}---\mintinline{python}{VanillaSaliency}
\cite{vanilla_saliency}, \mintinline{python}{GradCam} \cite{gradcam} and
\mintinline{python}{ScoreCam} \cite{scorecam}. In technical terms, these
visualisers work by adding noise to the input tensor and observing the gradient
of the class score. In the picture analogy, it would show us the pixels that
have the greatest effect on the classification result for an image. To
illustrate this point, Figure \ref{fig:sc_img} shows attention output from the
\mintinline{python}{ScoreCam} visualizer when used on images. When used on
poems, since each `pixel' is a syllable, it should show us specific syllables
that are salient to the given author's style. Since a syllable is embedded along
with all of its metrical metadata (Sec. \ref{sec:embed}, Fig.
\ref{fig:syl_transform}) the salient feature could be either sonic or metrical.

\begin{figure}
    \includegraphics[width=\linewidth]{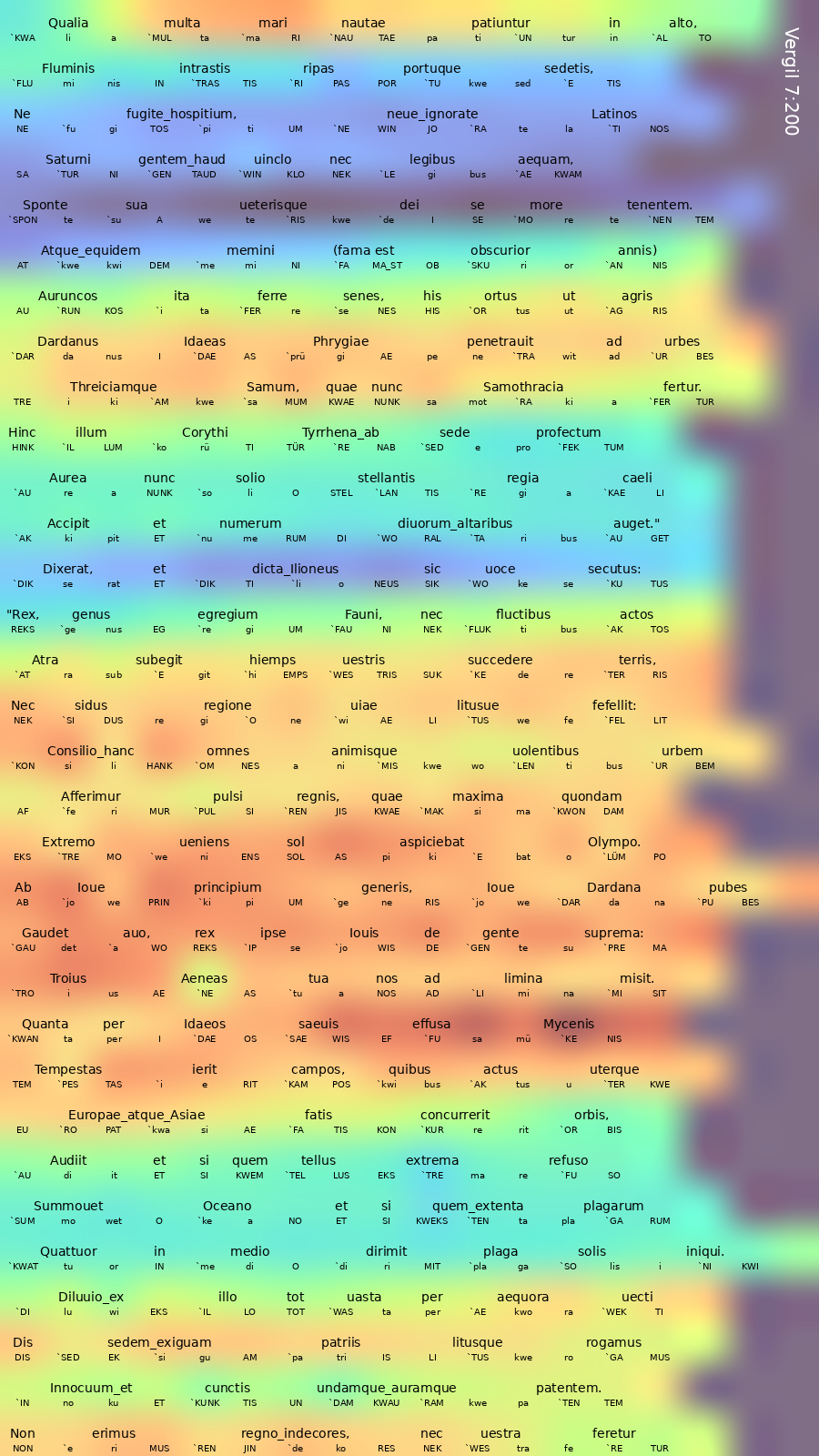}
    \caption{LSTM attention in detail, showing Vergil \emph{Aeneid}
        7.200--64. Some work-specific named entities appear to have high salience
        (\emph{Mycenis}, \emph{Dardana}, \emph{Troius}\textellipsis).}
    \label{fig:verg_lstm}
\end{figure}

It should also be mentioned that although \mintinline{python}{tf-keras-vis} is
aimed at visualising convolutional networks, the
\mintinline{python}{VanillaSaliency} tool can be used for LSTMs, with some
modifications. The visualisers work by perturbing real-valued inputs (each color
channel is usually mapped onto $[0,1]$). Instead of providing an integer vector
of encoded syllables to the LSTM model, the input vector can be embedded
manually, and then provided to \mintinline{python}{tf-keras-vis} as a tensor of
floats. To then visualise the output (Figs \ref{fig:thumbs32},
\ref{fig:verg_lstm}) it is simply reshaped to two dimensions so that each row
corresponds to one line of the original verse. Since LSTM networks have the
ability to learn sequences of syllables, they recognise things like named
entities that are unique to a certain work---this effect can be seen in Figure
\ref{fig:verg_lstm}. This could lead to an overfitting problem, but in
practice the issue did not seem to adversely affect accuracy on the unseen
verification data, even for samples that lacked any such `giveaway' names. In
theory this kind of sequence recognition is also possible with CNNs, but did not
seem to occur when examining the tests.

In the overall scheme of things, CNNs make better targets for attention
visualisation. LSTMs still suffer from the problem of \emph{vanishing gradients}
because the sample is a long sequence of syllables whose scores feed into each
other. This means that the most recent syllable seen by the network has
proportionally more effect on the gradient---this can be seen in Figure
\ref{fig:thumbs32} (looking closely, since the LSTM
model [\ref{app:lstm}] includes a \mintinline{python}{Bidirectional} layer we
can see a spurious attention hot-spot at the start as well as the end of each
sample). In contrast, since the CNNs were built to consider two-dimensional
data, according to the `sound picture' analogy, no region is inherently less
likely to contribute to the overall attention score. Another advantage is that
for CNNs there are additional visualizers available
(\mintinline{python}{GradCam}, \mintinline{python}{SmoothGrad}, etc.), which
focus not on the final dense layer, but on convolutional layers deeper in the
model (the final convolutional layer, by default). When understanding image
classification these more recent visualizers are less noisy; in the present set
of experiments it was not at all clear that this was the case.

\subsubsection{Problems at the dense layer}

\begin{figure*}
    \includegraphics[width=\textwidth]{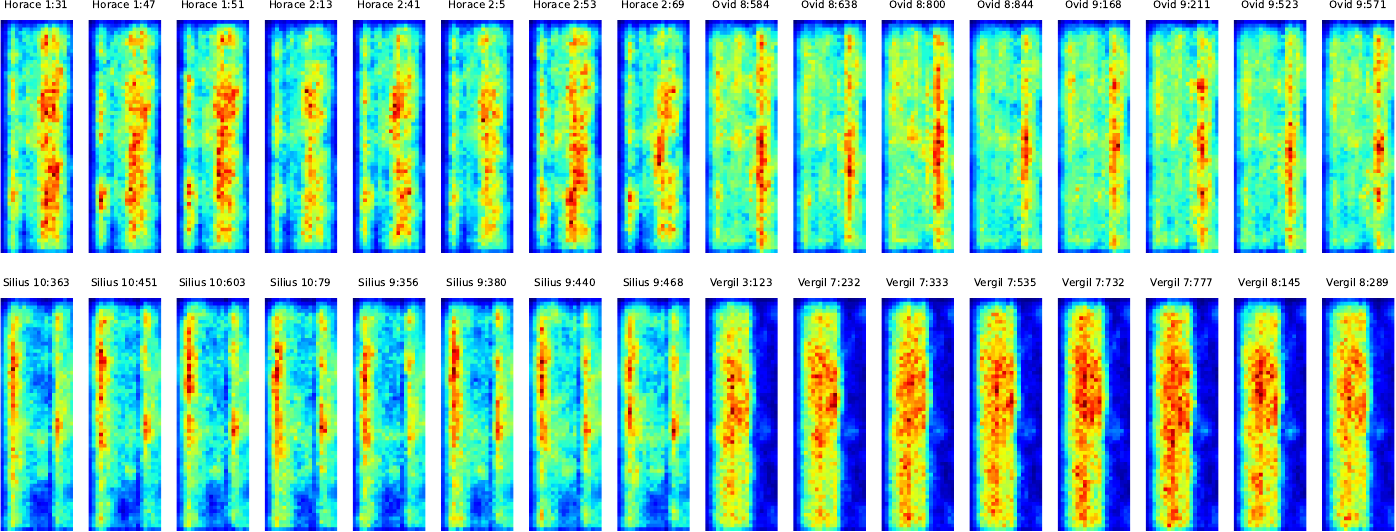}
    \caption{Comparing the \mintinline{python}{VanillaSaliency} view (CNN) of
        various passages from Horace, Ovid, Silius, and Vergil. Observe that the
        shapes of the heatmaps are mostly determined by the category, not by the
        content.}
    \label{fig:hosv-vs}
\end{figure*}

\begin{figure*}
    \includegraphics[width=\textwidth]{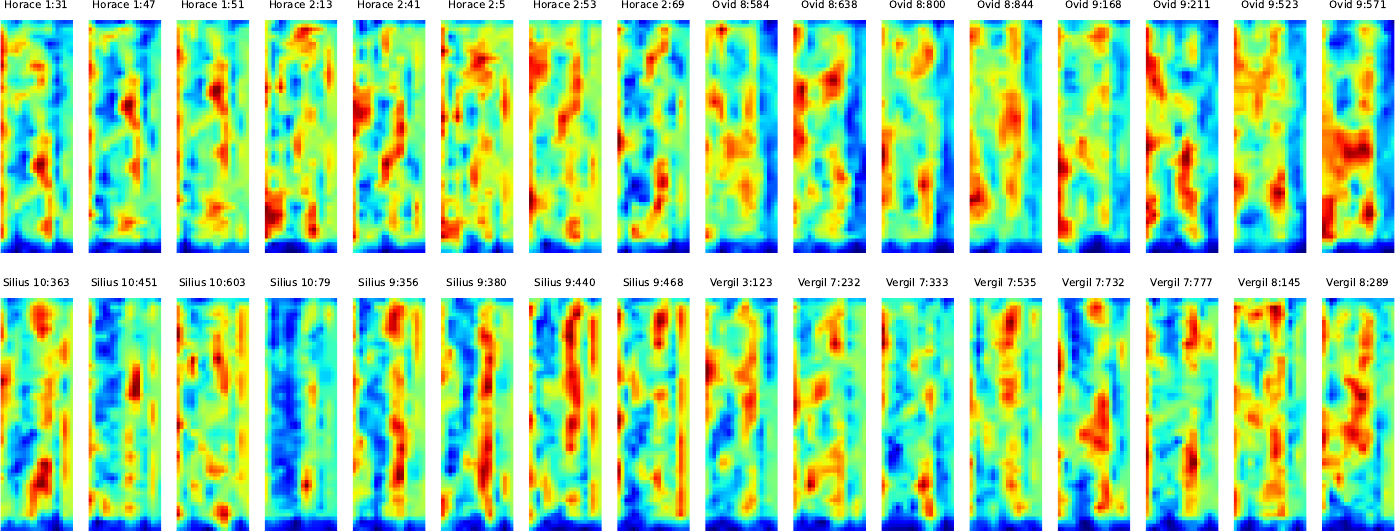}
    \caption{The \mintinline{python}{ScoreCam} view (CNN) of the same passages
        from Horace, Ovid, Silius, and Vergil (Fig. \ref{fig:hosv-vs}). The
        output now fully reflects the content. Some patterns are still similar
        (but shifted) because the passages overlap.}
    \label{fig:hosv-sc}
\end{figure*}

\begin{figure}
    \includegraphics[width=\linewidth]{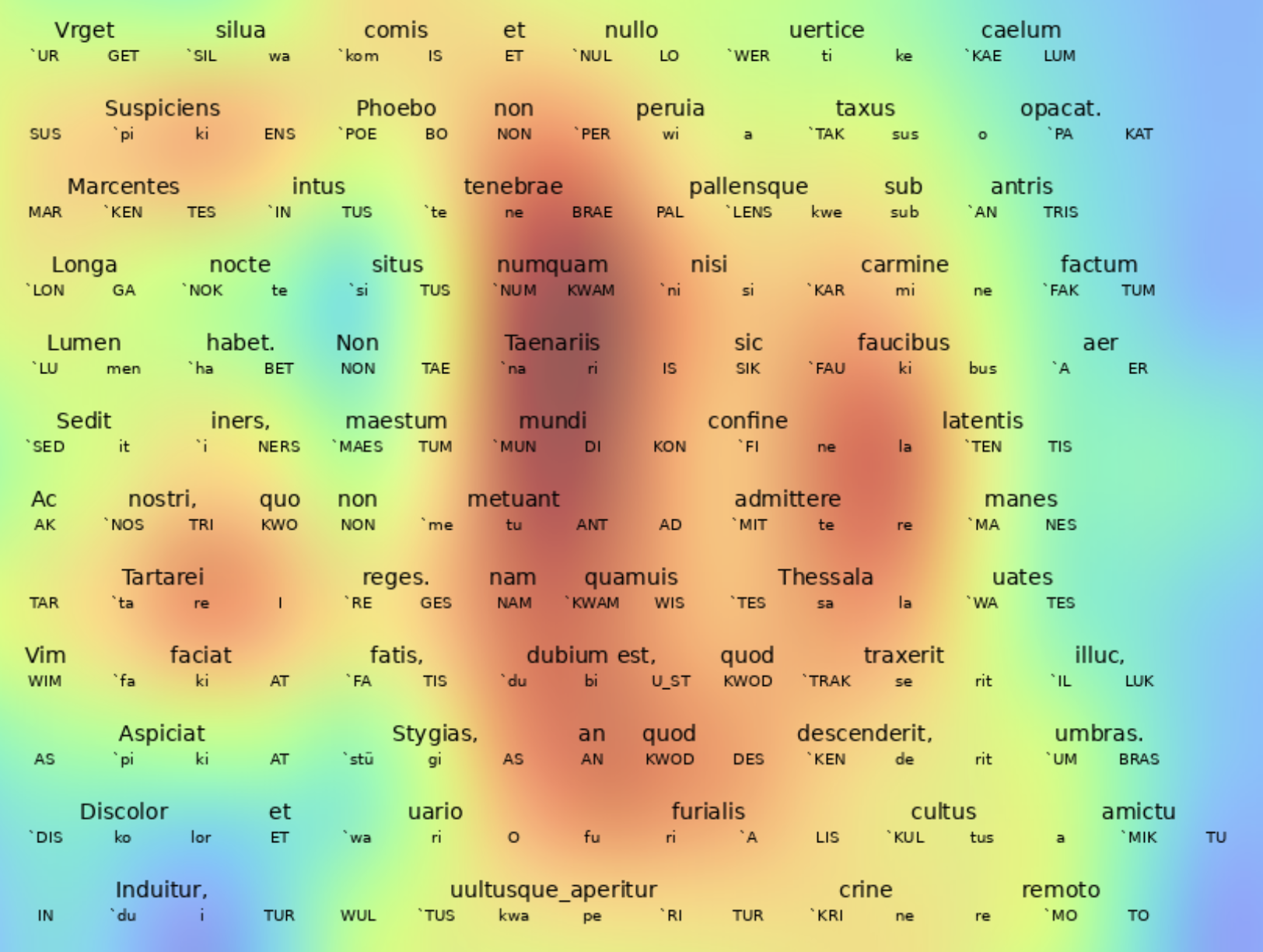}
    \caption{\mintinline{python}{ScoreCam} view (from the CNN model) of Lucan
        \emph{Pharsalia} 6.644--55. It is unclear what the convolutional filter
        finds interesting about the highlighted area.}
    \label{fig:lucan_scorecam}
\end{figure}

\begin{figure}
    \includegraphics[width=\linewidth]{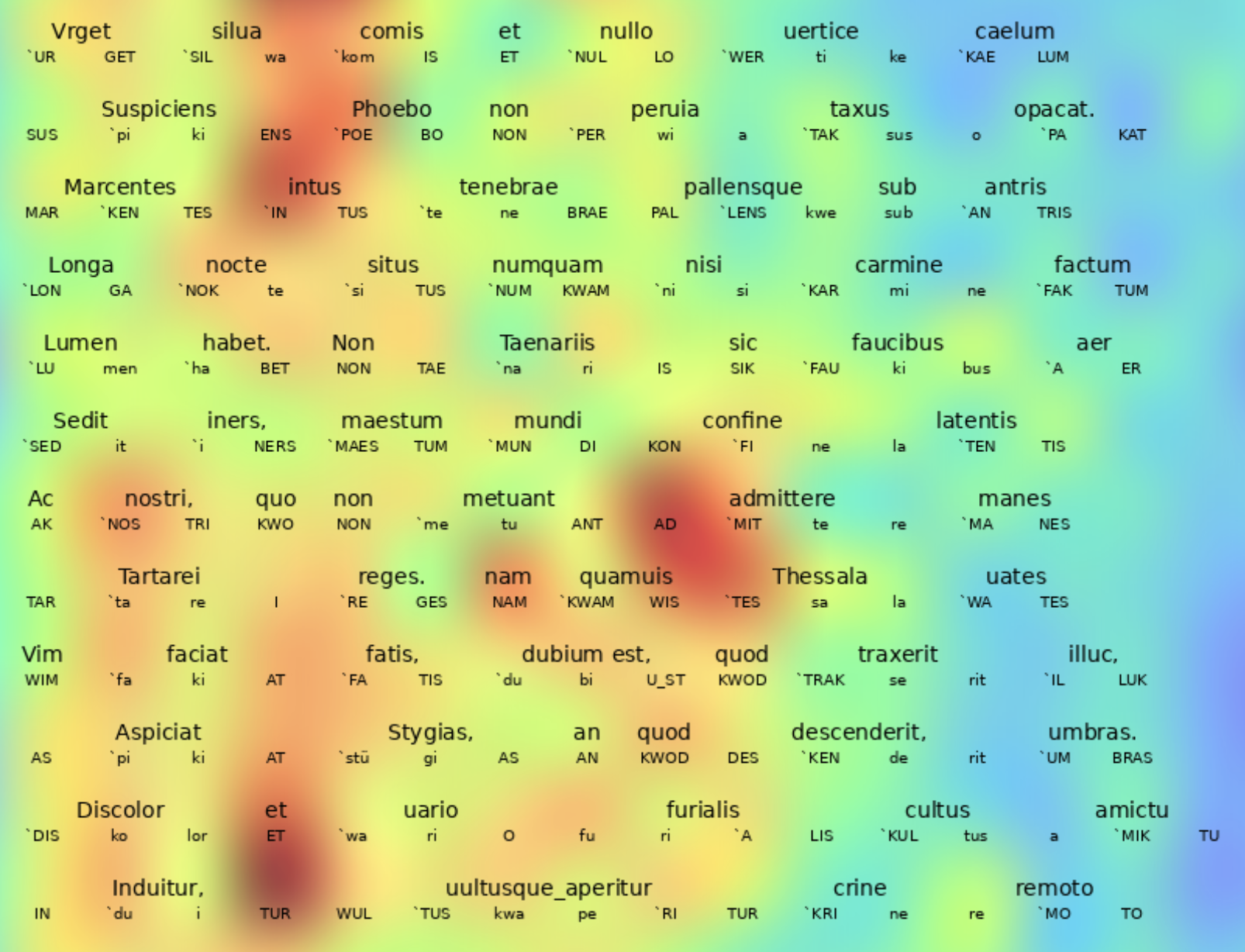}
    \caption{\mintinline{python}{VanillaSaliency} view (CNN) of the same passage
        (Fig. \ref{fig:lucan_scorecam})}
    \label{fig:lucan_vs}
\end{figure}

After a great deal of experimentation it became clear that visualising attention
at the dense (penultimate) layer was not effective in this application. By
looking at Figure \ref{fig:hosv-vs}, it can easily be seen that the
`attention' is mostly determined by the assigned class, and not by any
difference in the content of the passage (compare Figure \ref{fig:hosv-sc}, a
\mintinline{python}{ScoreCam} visualisation at the final convolutional layer).
Since the dense layer is fully connected, it appears that it `masks' areas of
the passage it thinks are the best places to look for stylistic markers for each
author; in the given example the attention for Vergil is mainly restricted to
the middle of the line, for Ovid to the end. Initially, this at least seemed to
be interesting poetic information---perhaps Ovid is distinguished by the way he
ends his lines. Unfortunately, this is not a valid conclusion. In Figure
\ref{fig:hosv-vs} we see that the areas examined for Vergil and Silius (who
have similar metrical styles) complement each other---Vergil in the middle of
the line, Silius at each end. It seems that essentially there is style available
everywhere, and the `masking' effect is (at least to some extent) a way of
partitioning the input domain that evolves to avoid class confusion. It is
possible that the problem might diminish if there were many more classes, but
for the present application the dense layer does not seem useful.

\subsubsection{Problems everywhere else as well}

As mentioned, CNNs offer additional possibilities for visualisation, since
attention can be visualised at the convolutional layers. As in the motivating
example shown as Figure \ref{fig:sc_img}, this might allow areas to be identified
as particularly typical of an author's style. So far, this has not been
effective. The caveats here are that each individual model behaves differently,
and so do the visualisations they produce at their convolutional layers. The
most relevant architectural concerns seem to be the number of convolutional
layers (the attention drawn from deeply layered models \emph{may} be more
difficult to understand) and the way \emph{pooling} is applied. Pooling is a
technique used in CNNs by which large images are reduced in size by combining
the filter outputs of adjacent pixels---a reductive technique that allows the
recognition of larger scale patterns, and also reduces the number of parameters
in the model. For image classification models, \mintinline{python}{MaxPooling}
seems to be the most common (c.f. the VGG16 design \cite{vgg16}) but in this
domain \mintinline{python}{AveragePooling} generally performed better. Despite
many experiments, no model design was discovered that produced consistently
useful attention output, although it often seemed tantalisingly close. In many
instances the attention output would seem to identify an interesting poetic
feature, but a more rigorous examination would show other instances where the
same feature was essentially ignored.

As just one example, we have included the attention visualisations from the CNN
model (\ref{app:cnn}) for the \mintinline{python}{VanillaSaliency} (Fig.
\ref{fig:lucan_vs}) and \mintinline{python}{ScoreCam} (Fig.
\ref{fig:lucan_scorecam}) visualisers from the \mintinline{python}{tf-keras-vis}
package. Although the section (a subsection of a 64-line passage) attracts
attention from the CNN, it is difficult for humans (or at least for this
particular human) to determine what is particularly Lucanian about this
passage. Three lines beginning with a double disyllable is unusual, but occurs
more frequently in Ovid and Valerius Flaccus than in Lucan. The unusual proper
noun Taenariis is not particularly characteristic of Lucan (in the test corpus
it occurs most frequently in Statius). The weak end rhyme ``admittere manes //
Thessala vates'' may be involved, but that is no help in explaining the other
highlighted areas. Essentially, although the CNN has made the correct
classification, visualising its attention does not appear to help us understand
its conception of poetic style. This is not to say that all is lost---it is
entirely possible that more effective techniques exist, or indeed that a
different, or more astute Latinist would see something here that I have missed.
To this end, a selection of the visualisation outputs have been collated as
convenient PDFs in the associated code repository (Sec. \ref{sec:data}).

\section{Future Work}

And so, given that the primary goal of the research (to understand how neural
networks understand style) was not achieved in any meaningful way, can anything
be salvaged? Yes, there are still some avenues that have not been fully
explored. First, the properties of the syllable embeddings have barely begun to
be examined, in particular whether any useful meaning lies in their linear
combinations (as with word embeddings) or in topological relationships that are
too complicated to survive the crude projection to two dimensions. Training
speed is hugely improved when using pre-trained embeddings, so it seems
reasonable to assume that this is a useful space to work in, and worth
understanding better.

Second, given the strong classification performance, it would be useful to
derive some measure of similarity or (even better) distinctiveness that could be
used in authorship attribution or verification problems. Some potential was
observed while iteratively reducing the sample sizes for the holdout data while
examining the confusion matrix---this gives us some coarse understanding about
which authors are easily confused with each other. By more carefully examining
the properties of misclassified samples, this can perhaps be extended to draw
conclusions about stylistic similarity and difference at the sample level. In
particular, since the samples overlap, one might assume the lines that differ
between a misclassified sample and a similar, correctly classified one are
stylistically causal. In terms of direct measures, the models can also be
configured to output classification logits, but it is by no means clear that the
`class probabilities' have anything to do with genuine posterior probability
estimates---the models are often too confident in their predictions. Another
early experiment that was abandoned was to take the final (pre-classification)
dense layer of a trained model and embed the samples as points in that high
dimensional space for clustering approaches. Again, this suffered from the
overconfidence problem. Very strong clustering was observed, but with no useful
topographical information for classification errors. Difficult decisions and
misclassifications were not placed `just over the line' near a border between
two clusters, but rather squarely in the middle of the mistaken class. However,
the idea of training a model to create a more useful sample space was not
explored. On balance, though, the approach feels baroque, and the statistical
properties of these spaces would be completely unknown, suggesting that existing
approaches that transform samples to feature probability vectors are probably
more sensible.

\FloatBarrier

\section{Conclusions}

There remains a vast gulf between quantitative \emph{description}---even
description in great detail---of the stylistic elements of poetry and the
semi-conscious process by which a human might perceive and describe a style as
`Ovidian' or `Lucretian'. This research attempted (and mostly failed) to bridge
that gap by examining the \emph{attention} of artificial neural networks. With
an accuracy of 97--98\%, neural network performance on classification tasks
(guessing the author of a sample) probably exceeds that of human experts.%
\footnote{
    This is effectively impossible to test. The classical corpus is small, so
    any expert sufficiently versed in, say, Ovidian style will certainly have read
    the `unseen' passages by which we seek to evaluate them.}
However, if the NNs meaningfully `understand' poetic style, then the methods
used here do not bring us much closer to knowing how they do it. It is possible
that these approaches might succeed elsewhere---the classical Latin corpus is
small in comparison to modern languages, and the authors few. Perhaps better
methods would serve to extract this understanding: elicitation, generative
models or even NLP approaches by which models might paraphrase their
understanding in human language. Or, it is possible that the phenomenon we
experience as `poetic feel' is an emergent property of the human brain requiring
vastly more complicated models to emulate---a question for philosophy or
neuroscience. In any case, scientific advance is built on negative results as
well as positive ones. Whether future research improves these methods to a point
where they succeed, or whether it simply saves someone from retracing these
steps, it is hoped that these novel experiments will advance the field.

\section{Availability of Data and Code}\label{sec:data}

The preprint may be found at \url{https://github.com/bnagy/fail-paper}. All code
and data is available under CC-BY, except where restricted by upstream licenses.
The code repository includes all of the source data, the saved models, and
Python notebooks to replicate all figures contained in this paper, as well as
various supplemental figures and explanations.

\begin{acknowledgments}
    In early 2022, I was an anonymous reviewer of an interesting paper
    \cite{yilmaz_scheffler_22} which attempted authorship attribution of a large
    corpus of modern English song lyrics using CNNs. The early progress made in
    that paper towards examining rhyme with 1D convolutional networks, as well
    as inferring significance with occlusion maps was an important inspiration
    for this research (although the methods here are unrelated except in
    concept). This work was supported by Polish Academy of Sciences Grant
    2020/39 / O / HS2 / 02931.
\end{acknowledgments}

\bibliography{fail}

\appendix
\onecolumn
\section{Model Configurations}
\subsection{LSTM}\label{app:lstm}
\begin{minted}{python}
lstm = tf.keras.Sequential(
    [
        tf.keras.layers.Embedding(
            input_dim=len(encoder.get_vocabulary()),
            output_dim=32,
            mask_zero=True,
            name='embed'
        ),
        tf.keras.layers.Dropout(0.2),
        tf.keras.layers.Bidirectional(
            tf.keras.layers.LSTM(32,return_sequences=True),
            name='lstm1'
        ),
        tf.keras.layers.Dropout(0.2),
        tf.keras.layers.LSTM(32, name='lstm2'),
        tf.keras.layers.BatchNormalization(),
        tf.keras.layers.Dropout(0.2),
        tf.keras.layers.Dense(64, activation="relu", name='dense'),
        tf.keras.layers.Dropout(0.2),
        tf.keras.layers.Dense(n_classes, activation="softmax", name='classify'),
    ]
)

lstm.compile(
    loss="sparse_categorical_crossentropy",
    optimizer=tf.keras.optimizers.Adam(learning_rate=0.0005),
    metrics=["accuracy"],
)
\end{minted}

\subsection{CNN}\label{app:cnn}
\begin{minted}{python}
cnn = tf.keras.models.Sequential([
  # Take a 64 line sample, padded to 20 syllables per line and flatten it
  tf.keras.layers.Flatten(),
  # then embed
  tf.keras.layers.Embedding(
      name='embed',
      input_dim=len(encoder.get_vocabulary()),
      output_dim=32,
  ),
  # then reshape it to 64x20 again, with 32 bit embedding per 'pixel'
  tf.keras.layers.Reshape((64,20,32)),
  tf.keras.layers.Dropout(0.25),
  tf.keras.layers.Conv2D(
    24,
    # wider than long to de-emphasise distant vertical relationships
    kernel_size=(4,2), 
    activation='relu',
    padding='same',
    name='conv1'
  ),
  tf.keras.layers.AveragePooling2D(pool_size=(2, 2), strides=(2,2)),
  tf.keras.layers.BatchNormalization(name='norm1'),
  tf.keras.layers.Dropout(0.25),
  tf.keras.layers.Conv2D(
    48,
    kernel_size=(4,2),
    activation='relu',
    padding='same',
    name='conv2'
  ),
  tf.keras.layers.AveragePooling2D(pool_size=(2, 2), strides=(2,2)),
  tf.keras.layers.BatchNormalization(name='norm2'),
  tf.keras.layers.Flatten(),
  tf.keras.layers.Dropout(0.5),
  tf.keras.layers.Dense(64, name='dense1'),
  tf.keras.layers.Dense(64, name='dense2'),
  tf.keras.layers.Dropout(0.5),
  tf.keras.layers.Dense(n_classes, activation='softmax', name='classify')
])
cnn.compile(
    loss='sparse_categorical_crossentropy',
    optimizer=tf.keras.optimizers.Adam(learning_rate=0.0001),
    metrics=['accuracy'],
)
cnn.build((None,64,20))
\end{minted}
\end{document}